\documentclass[11pt,a4paper]{article}
\usepackage[hyperref]{naaclhlt2019}
\usepackage{times}
\usepackage{latexsym}

\usepackage{hyperref}
\usepackage{url}
\usepackage{booktabs}
\usepackage{multirow}
\usepackage{graphicx}
\usepackage{amsmath}
\usepackage{mathtools}
\usepackage{amssymb}
\usepackage{amsthm}
\usepackage{amsfonts}
\usepackage{csquotes}
\usepackage{breqn}
\usepackage{subfig}

\usepackage{enumitem}
\usepackage{paralist}

\usepackage[normalem]{ulem}
\useunder{\uline}{\ul}{}

\usepackage{url}

\aclfinalcopy % Uncomment this line for the final submission
\def\aclpaperid{N/A} %  Enter the acl Paper ID here

\title{Towards Coherent and Cohesive Long-form Text Generation}

\author{Woon Sang Cho$^{\star}$\quad Pengchuan Zhang$^{\dagger}$\quad Yizhe Zhang$^{\dagger}$\quad Xiujun Li$^{\dagger}$\quad \\
\textbf{Michel Galley$^{\dagger}$\quad Chris Brockett$^{\dagger}$\quad Mengdi Wang$^{\star}$\quad Jianfeng Gao$^{\dagger}$}  \\
  $^{\star}$Princeton University \\
  $^{\dagger}$Microsoft Research AI \\
  {\tt $^\star$\{woonsang,mengdiw\}@princeton.edu} \\
  {\tt $^\dagger$\{penzhan,yizzhang,xiul,mgalley,chrisbkt,jfgao\}@microsoft.com}
}

\date{}

\begin{document}

\maketitle

\begin{abstract}
Generating coherent and cohesive long-form texts is a challenging task. Previous works relied on large amounts of human-generated texts to train neural language models. However, few attempted to explicitly improve neural language models from the perspectives of coherence and cohesion. In this work, we propose a new neural language model that is equipped with two neural discriminators which provide feedback signals at the levels of sentence (cohesion) and paragraph (coherence). Our model is trained using a simple yet efficient variant of policy gradient, called \textit{negative-critical sequence training}, which is proposed to eliminate the need of training a separate critic for estimating \textit{baseline}. Results demonstrate the effectiveness of our approach, showing improvements over the strong baseline -- recurrent attention-based bidirectional MLE-trained neural language model.

% Generating coherent and cohesive long-form texts is a challenging problem in natural language generation. Previous works relied on a large amount of human-generated texts to train neural language models; however, few attempted to explicitly model the desired linguistic properties of natural language text, such as coherence and cohesion using neural networks. In this work, we train two expert discriminators for coherence and cohesion to provide hierarchical feedback for text generation. We also propose a simple variant of policy gradient called \textit{negative-critical sequence training}, in which the reward \textit{baseline} is constructed from randomly generated negative samples. Results from automatic and human evaluation demonstrate the effectiveness of our approach, showing improvements over the strong baseline -- attention-based bidirectional MLE-trained neural language model.
\end{abstract}

\section{Introduction}
The terms \textit{coherence} and \textit{cohesion} in linguistics are commonly defined as follows \citep{williams1995style}.

\begin{compactitem}
    \item \textit{Cohesion}: sentence pairs fitting together the way two pieces of a jigsaw puzzle do.
    \item \textit{Coherence}: what all the sentences in a piece of writing add up to, the way all the pieces in a puzzle add up to the picture on the box.
\end{compactitem}

In layman's terms, \textit{cohesion} indicates that two consecutive sentences are \textit{locally} well-connected, and \textit{coherence} indicates that multiple sentences \textit{globally} hold together. 

Generating cohesive and coherent natural language texts that span multiple sentences is a challenging task for two principal reasons. First, there is no formal specification of cross-sentence linguistic properties, such as coherence and cohesion of a text. Secondly, there is no widely accepted model to measure the two properties. 

Most state-of-the-art neural approaches to natural language generation rely on a large amount of human-generated text to train language models \citep{graves2013generating,cho-al-emnlp14,NIPS2014_5346}. Although these models can generate sentences that, if judged individually, are similar to human-generated ones, they often fail to capture the local and global dependencies among sentences, resulting in a text that is neither coherent nor cohesive. For example, neural language models based on Recurrent Neural Networks (RNNs) are widely applied to response generation for dialogue \citep{vinyals2015neural, shang2015neural, sordoni2015neural, li2015diversity}. Although the responses by themselves look reasonable, they are detached from the whole dialogue session. See \citet{gaosurvey} for a comprehensive survey. 

In this paper, we address the challenge in a principled manner, employing a pair of discriminators to score whether and to what extent a text is coherent or cohesive. The coherence discriminator measures the compatibility among all sentences in a paragraph. The cohesion discriminator measures the compatibility of each pair of consecutive sentences. These models, given a conditional input text and multiple candidate output texts, are learned to score the candidates with respect to the criterion. The scores are used as reward signals to train an RNN-based language model to generate (more) coherent and cohesive texts. 

\paragraph{Contributions.} Our main contributions are: ({\bf 1}) we propose two neural discriminators for modeling coherence and cohesion of a text for long-form text generation; %To the best of our knowledge, this paper is the first to explicitly capture cross-sentence linguistic properties (coherence and cohesion) in a fully end-to-end neural long-form text generation setting; 
({\bf 2}) we present a simple yet effective training mechanism to encode these linguistic properties; ({\bf 3}) we propose \emph{negative-critical sequence training}, a policy gradient method that uses negative samples to estimate its reward \emph{baseline} and therefore eliminates the need for a separate critic function; and ({\bf 4}) we develop a new neural language model that generates more coherent and cohesive long-form texts, and empirically validate its effectiveness using the TripAdvisor and Yelp English reviews datasets.

% This paper proposes a neural approach to explicitly modeling cross-sentence linguistic properties, coherence and cohesion, for long-form text generation. Despite the encouraging initial results, we only scratched the surface of the problem. The proposed method is yet to be significantly improved to meet the ultimate goal of generating meaningful and logical long-form texts. \JG{Move this paragraph to the Conclusion section.} 

%We cast the text generation as a reinforcement learning (RL) problem and review recent work in Section~\ref{related}, and detail our approach in Section~\ref{model}.

\section{Related work} 
\label{related}

\paragraph{Coherence and cohesion.} 
Coherence and cohesion have been extensively studied in the computational linguistics community, particularly in the `pre-deep-learning' era. Lack of formal specifications for coherence and cohesion \citep{mani1998using}, resulted in many different formalisms, such as Rhetorical Structure Theory \citep{mann1988rhetorical}, and other forms of coherence and cohesion relations and their quantification \citep{edmundson1969new,halliday1996cohesion,hobbs1985coherence,mckeown1985discourse,cohen1985speech,hovy1988planning,liddy1991discourse,hovy1991approaches,mani1998using,cristea1998veins,barzilay2008modeling,van2013news}. This list is not exhaustive. However, prior work jointly exploring coherence and cohesion using neural models in the context of long-form text generation has not come to our attention.
%However, modeling coherence and cohesion of a text using neural language models have not been previously explored. 

\paragraph{Reinforcement learning for text generation.}
%Word sequence generation in a reinforcement learning framework.}
The text generation task can be framed as a reinforcement learning (RL) problem \cite{DBLP:journals/corr/abs-0907-0786}, in which the generator $G$ is acting as a \textit{policy} $\pi$, with parameters $\theta_{\pi}$, and each generated word at time $t$, $w_{t}$, can be viewed as an action to be chosen by the policy from a large discrete space, or vocabulary, conditioned on state $s_{t-1}=w_{\leq t-1}$.%, which encodes the previously generated text sequence.

Let $r_{t}$ be the reward for a partially generated text sequence $w_{\leq t}$. We define the long-term expected reward $\mathcal{J}(\pi) = \mathbb{E}_{s_{0} \sim q, \pi} \big[\sum_{t=1}^{\infty}\gamma^{t-1}r_{t} \big]$, where $q$ is the initial distribution of conditional input texts. Following \citet{Sutton:1999:PGM:3009657.3009806}, the gradient of $\mathcal{J}$ with respect to $\theta_{\pi}$ is
% \begin{equation} \label{policygradient}
\[
\resizebox{\columnwidth}{!}{$
\nabla_{\theta_{\pi}}\mathcal{J}=~
\mathbb{E}_{s \sim \rho^{\pi}, a \sim \pi(\cdot \vert s)} \big[ Q^{\pi}(s,a) \nabla_{\theta_{\pi}} \log \pi_{\theta_{\pi}}(a\vert s) \big]$}
\]
% \end{equation}
where $\rho^{\pi}$ is the stationary distribution and $Q^{\pi}(s,a)$ is the expected return from state $s$ and taking action $a$, both following policy $\pi$. For brevity, we omit the derivation. In this work, we formulate text generation as an episodic RL problem with episode length $L$, rewards $r_{L}$ being available only at the end of episode and $\gamma=1$.

There are many works on training neural language models using rewards, such as \citet{DBLP:journals/corr/RanzatoCAZ15} and \citet{DBLP:journals/corr/PaulusXS17}. These works directly optimize for specific metrics, such as BLEU \citep{Papineni:2002:BMA:1073083.1073135} or ROUGE \citep{Lin:2003:AES:1073445.1073465}, 
%or METEOR \citep{Lavie:2007:MAM:1626355.1626389} to list a few, 
using REINFORCE \citep{Williams:1992:SSG:139611.139614}. However, %it is well-known that
these metrics do not give a complete picture of the text generation quality. Only recently have there been efforts to provide more relevant objectives, such as consistency and repetition in a text \citep{li2015diversity,li2016persona,holtzman2018l2w}. But these works use the objectives to re-rank candidate outputs, not to reward or penalize them. %when they are generated in the first place. 
\citet{li2016deep} constructed a set of reward models for the dialogue task, such as information flow and semantic coherence, to tune the generator, yet they do not provide an ablation study on the relative contribution of these reward models individually. It is not clear that these reward models can be generalized to other tasks, in particular, long-form text generation tasks. 

The most relevant to our work is \citet{bosselut18discourse}, which promotes text generation in the correct order, and discourages in its reverse order using rewards. However, this may not be sufficient in capturing coherence since there are many negative orderings given a paragraph. From this pool, we assess the relative quality of generations. Furthermore, we model cohesion between consecutive sentence pairs using word-level features.

\paragraph{GANs for text generation.} 
Another line of research involves the use of Generative Adversarial Networks (GANs) \citep{NIPS2014_5423} to incorporate feedback signals for text generation \citep{Yu2017SeqGANSG,lin2017adversarial,zhang2017adversarial,guo2017long,46657,zhang2018generating}. The discriminators in these works are trained to distinguish real texts from generated ones, operating as a black-box than providing feedback on linguistic aspects. % of the texts. %In fact,  
\citet{yang2018unsupervised} partially addressed this issue by using a trained language model as the discriminator. Although the discriminator provides a fine-grained feedback at the word level, it does not model linguistic properties, such as cohesion and coherence. 

% \yz{probably mention some text generation papers using hierarchical structures here. Especially those with sentence-level and word-level hierarchy.
% Consider revise based on below text:
% Latent variable based hierarchical encoder-decoder framework was employed by several existing works, including paragraph reconstruction \citep{li2015hierarchical}, conversation modelling~\citep{Serban2017AHL}.
% \citet{yang2016hierarchical} employs hierarchical LSTM encoders at the different hierarchy to learn document representations. 
% }

%Most state of the art text generators are not sufficient to generate a cohesive and coherent long-form text that span multiple sentences, 
%Many generator models, when faced with a long-form text generation task that span multiple sentences, are by no means perfect and often exhibit some critical errors, such as a breakdown of local connections between consecutive sentences (cohesion), let alone globally solid intention (coherence). 

Many text generator models are inadequate for generating a cohesive and coherent long-form text that span multiple sentences.
As a result, human readers can easily distinguish the generated texts from real ones. 
In this paper, we argue that the primary reason is the lack of an effective mechanism to measure and control for the local and global consistency in model-generated texts.
%in the generation setting.
%The method we propose in the next section is intended to address the problem. % This motivates our work.

\section{Coherence and Cohesion Models}
\label{model}

We assume that global coherence of a text depends to a large degree upon
%soundness is governed by 
how its individual sentences with different meanings are organized. %to form a single paragraph. 
Therefore, we focus our evaluation of coherence solely based on the sentence-level features. If the sentences are not organized properly, the intention %message 
of the paragraph as a whole is obscure, regardless of seamless local connectivity between consecutive sentences. 

This is not to say that local connections between any two neighboring sentences can be overlooked. One can easily distinguish a generated sentence from a real one by judging whether it is \emph{semantically cohesive} with its neighboring sentences. 
% sound sentence, not to speak of grammar. 

We strive to embody these two different yet important concepts by developing coherence and cohesion discriminators, operating on the sentence level and word level, respectively. 
% We call the sentence-level discriminator the \textit{coherence} discriminator $D_{\text{coherence}}$, and the word-level discriminator the \textit{cohesion} discriminator $D_{\text{cohesion}}$. 
Our design of these two discriminators is inspired by the Deep Structured Semantic Model (DSSM) which was originally developed to measure the semantic similarity between two texts \citep{Huang2013LearningDS,gao2014modeling,hamiddssm,xu2017attngan}. In this study, we extend `semantic similarity' to coherence and cohesion in a long-form text. 

\subsection{Coherence discriminator: \texorpdfstring{$D_{\text{coherence}}$}{Lg}}
\label{subsec:coherence}
The coherence discriminator models the coherence score, which measures how likely two text chunks add up to a single coherent paragraph. Let $S \coloneqq \left[ s_{1}, s_{2},..., s_{n} \right]$ be the source text chunk that consists of $n$ sentences, $T \coloneqq \left[ t_{1}, t_{2},..., t_{m} \right]$ be the \emph{real} target text chunk that consists of $m$ sentences, and $\widetilde{T} \coloneqq \left[ \widetilde{t}_{1}, \widetilde{t}_{2},..., \widetilde{t}_{\widetilde{m}} \right]$ be the \emph{artificially constructed incoherent} target text chunk that consists of $\widetilde{m}$ sentences. $D_{\text{coherence}}$ is designed to distinguish a positive (coherent) pair $\left(S,T \right)$ from a negative (incoherent) pair $(S,\widetilde{T})$ by assigning different scores, i.e., $D_{\text{coherence}}(S, T) > D_{\text{coherence}}(S, \widetilde{T} )$.
%$\text{Score}_{\text{coherence}}(S,T) > \text{Score}_{\text{coherence}}(S,\widetilde{T})$.

\begin{figure}[ht!]
% \begin{center}
\centering
% \resizebox{\textwidth}{!}{
\includegraphics[width=1.00\linewidth]{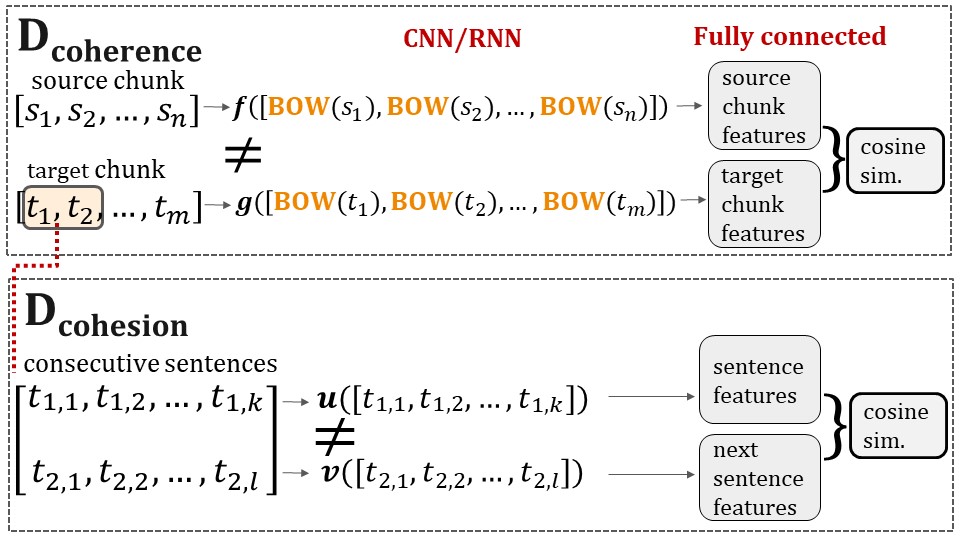} %0.91
% }
% \end{center}
\caption{Illustration of coherence and cohesion discriminators. $D_{\text{coherence}}$ takes in bag-of-words sentence embeddings as inputs, and $D_{\text{cohesion}}$ takes in the raw word embeddings of consecutive sentences as inputs. The source encoder $f$ (or $u$) is different from the target encoder $g$ (or $v$).}
\label{dssm}
\end{figure}

\paragraph{Model architecture.} 
The model takes a form of dual encoder. 
Given source text chunk $S$ and target text chunk $T$, the coherence discriminator $D_{\text{coherence}}$ computes the coherence score in three steps, as illustrated in Figure~\ref{dssm} (upper).
First, each sentence is encoded by the bag-of-words (BOW) embedding, i.e., the average of its word vectors from a pre-trained word embedding \citep{pennington2014glove}.   
Secondly, an encoder which can be implemented using a convolutional neural network (CNN)\footnote{We explored with deeper networks. However, the performance difference was marginal. For simplicity, we decided to use a 1-layer convolutional network architecture \citep{Kim2014ConvolutionalNN,Collobert:2011:NLP:1953048.2078186}.} or RNN\footnote{For clarity in our model description, we omit RNN hereafter. We present results using both CNN and RNN encoders in Table~\ref{retrieval}.}, denoted as $f$, takes as input the BOW vectors of the source text chunk $S$ and encodes it into a single vector $f(S)$. Similarly, $g$ encodes the target text chunk $T$ into $g(T)$. The two encoders $f(\mathord{\cdot})$ and $g(\mathord{\cdot})$ share the same architecture but do \textit{not} share parameters, i.e.,  $\theta_f \neq \theta_g$, and thus $D_{\text{coherence}}(S, T)$ is \textit{not} symmetric. 
Thirdly, $D_{\text{coherence}}(S, T)$ is computed as the cosine similarity of the two vectors $f(S)$ and $g(T)$. The score is a real value between $-1$ and 1, where 1 indicates maximal coherence, and $-1$ minimal coherence.

Note that we use the simple BOW vectors to encode sentences in the coherence discriminator, which is different from the CNN sentence embedding scheme in the cohesion discriminator that we introduce in Section \ref{subsec:cohesion}. Although the BOW vector ignores the word-order information in the sentence, it is empirically shown to be effective in preserving the high-level semantic information in the sentences and achieves success in sentence similarity and entailment tasks \citep{wieting2016iclr, arora2017asimple}. Because high-level semantic information of sentences is sufficient to determine whether a paragraph is coherent, we choose to use BOW vectors to encode sentences in $D_{\text{coherence}}$. 

% Since we focus solely on the sentence-level features, we view a text chunk as a sequence of sentences, and view each sentence as a bag-of-words. Therefore, we represent each word using its pre-trained word embedding vector \citep{pennington2014glove}
% and represent each sentence using a vector that takes the average of its word embedding vectors. %A text chunk is then represented as a sequence of sentence vectors that are fed to either 

%The parameters of $f(\mathord{\cdot})$ and $g(\mathord{\cdot})$ are optimized in such a way that a real pair scores higher than a synthetic pair, i.e., $\Delta(\theta_f,\theta_g) > 0$, where 
% \begin{equation*}
% % \label{regular_margin}
% \begin{split}
% \Delta(\theta_f,\theta_g) \coloneqq D_{\text{coherence}}\Big(f_{\theta_f}(S),g_{\theta_g}(T)\Big) \\ -  D_{\text{coherence}}\Big(f_{\theta_f}(S),g_{\theta_g}(\widetilde{T})\Big)
% \end{split}
% \end{equation*}
% , the task of optimizing $D_{\text{coherence}}$ can be stated as follows. Given a set of training samples of the form $((S,T), (S,\widetilde{T}))^{(i)}, i=1...M$, we optimize parameters $(\theta_f,\theta_g)$ by minimizing the pairwise margin ranking loss on training data defined as
% % \[
% % \frac{1}{M} \sum_{i=1}^M L(\Delta(\theta_f,\theta_g)^{(i)})
% % \]
% \[
% L(\mathord{\cdot}) \coloneqq \frac{1}{M} \sum_{i=1}^M \max(0, \delta-\Delta(\theta_f,\theta_g)^{(i)})
% \]
% where $L(\mathord{\cdot})$ is a loss function, differentiable w.r.t. $(\theta_f,\theta_g)$.

The parameters of $D_{\text{coherence}}$, $\theta_f$ and $\theta_g$ are optimized using a pairwise ranking loss. To this end, we need both positive and negative pairs. While the positive (coherent) pairs come from the training data, negative (incoherent) pairs need to be artificially constructed.
The next section describes the way these negative pairs are generated.

% We artificially construct negative (incoherent) pairs from mismatched ones in a minibatch (rotation) and reordering sentences in the target chunk (shuffling). 

%In this subsection, we will describe in turn how we construct the pairwise negative training samples of the form $\{(S,\widetilde{T_{j}})\}_{j=1}^{2B-1}$ for each positive sample $(S,T)$, where $B$ is the batch size, and the form of the loss function $L(\mathord{\cdot})$. 
%$\Big\{(S,T), \{(S,\widetilde{T})\}^{(j)} \Big\}^{(i)}, \; i=1,\dots,M$,
% $\Big\{(S,T), \{(S_{i},\widetilde{T_{ij}})\}_{j=1}^{2B-1} \Big\}_{i=1}^{B}$
% How do we construct these list of incoherent target chunk $\widetilde{T_{j}}$, given the source chunk $S$? We assume the $T$ that follows an $S$ in the data is a positive target sample, or the correct item to retrieve. 

%$\{(S,\widetilde{T_{j}})\}_{j=1}^{2B-1}$
\paragraph{Constructing negative (incoherent) pairs.} 
Given a training minibatch $\{(S_i,T_i)\}_{i=1}^B$, we construct $2*B-1$ negative pairs $\{(S_i,\widetilde{T_{i,j}})\}_{j=1}^{2B-1}$ for \textit{every} positive pair $(S_i, T_i)$ using three different methods, inspired by \citet{wieting2016iclr}. For notation simplicity, we omit the minibatch index $i$ in the rest of this section. For each positive pair $(S, T)$ in the minibatch:
% $\{(S_{i},T_{i}),(S_{i},\widetilde{T_{ij}})\}_{j=1,\dots,2B-1}^{i=1,\dots,B}$
\begin{compactitem}
\item We rotate $T$ with $S$ fixed, and thus obtain all $B-1$ mismatched pairs $\{(S,\widetilde{T_{j}})\}_{j=1}^{B-1}$ as negative pairs. 
% $B-1$ negative samples for a single $S$
\item We shuffle the sentence order in $T$ once, known as a derangement, to break its coherence. This yields one negative pair ($S,\widetilde{T}$).
\item We combine the previous two methods, that is, we rotate $T$ in the minibatch and shuffle sentences within the target chunk, yielding another $B-1$ negative pairs $\{(S,\widetilde{T_{j}})\}_{j=1}^{B-1}$. 
\end{compactitem}

These $2B-1$ negative pairs and a single positive pair, in total, pose a challenge for the discriminator in learning to retrieve the correct pair.

\paragraph{Training using a pairwise ranking loss.}

The parameters of $f(\mathord{\cdot})$ and $g(\mathord{\cdot})$ are optimized in such a way that a positive pair scores higher than its negative pairs, i.e., $D_{\text{coherence}}(S, T) >  D_{\text{coherence}}(S, \widetilde{T}_j)$ for any $j$. 
To achieve this, we propose to minimize the following pairwise ranking loss~\cite{gong2013deep} with margin $\delta$:
\begin{equation}\label{coherence_loss_function} 
\resizebox{0.89\columnwidth}{!}{$
\begin{aligned}
	&L_{\text{coherence}}(\theta_f,\theta_g) \coloneqq \max\Big(0, \delta - D_{\text{coherence}}(S, T) \\
	&+ \text{AVG}^{\lambda} \left( \{D_{\text{coherence}} (S, \widetilde{T}_{j})\}_{j=1}^{2B-1} \right) \Big).
\end{aligned}$}
\end{equation}
where $\text{AVG}^{\lambda}(\{x_j\}_{j=1}^N) = \sum\nolimits^{N}_{j=1}w_{j}x_{j}$ and $w_{j}=e^{\lambda x_{j}}/\sum_{k}e^{\lambda x_{k}}$.  

Notice that $\text{AVG}^{\lambda}$ is the \textit{mean} operator when $\lambda=0$ and approaches the \textit{max} operator when $\lambda \rightarrow \infty$. These two extreme cases correspond to ranking against the average of all negative pairs and ranking against the single most challenging negative pair, respectively. Empirically, training the models using the \textit{weighted} average ($0 < \lambda \ll \infty $), which assigns larger weights to more challenging negative pairs, stabilizes the training and expedites the convergence. 

\subsection{Cohesion discriminator: \texorpdfstring{$D_{\text{cohesion}}$}{Lg}}
\label{subsec:cohesion}
% The cohesion discriminator, as opposed to the coherence discriminator, pays attention only to low-level features to capture local connectivity between arbitrary two consecutive sentences. For simplicity, $D_{\text{cohesion}}$ is similar to $D_{\text{coherence}}$, except that its architecture, input, and negative sample construction are modified to encode cohesion between any pair of sentences on the word-level. 

The cohesion discriminator models the cohesion score, which measures how likely two sentences form a cohesive pair of consecutive sentences. 
Let $s_{k}\coloneqq \left[ s_{k}^{1}, s_{k}^{2},..., s_{k}^{n} \right]$ be the $k^{\text{th}}$ sentence that consists of $n$ words, $s_{k+1} \coloneqq \left[ s_{k+1}^{1}, s_{k+1}^{2},..., s_{k+1}^{m} \right]$ be the \emph{real} next sentence that consists of $m$ words, and $\widetilde{s}_{k+1} \coloneqq \left[ \widetilde{s}_{k+1}^{1}, \widetilde{s}_{k+1}^{2},..., \widetilde{s}_{k+1}^{\widetilde{m}} \right]$ be the \emph{artificially constructed incohesive} next sentence that consists of $\widetilde{m}$ words. $D_{\text{cohesion}}$ is designed to distinguish a positive (cohesive) pair $\left(s_{k},s_{k+1} \right)$ from a negative (incohesive) pair $(s_{k},\widetilde{s}_{k+1})$ by assigning them with different scores, i.e., $D_{\text{cohesion}}(s_{k},s_{k+1}) > D_{\text{cohesion}}(s_{k},\widetilde{s}_{k+1})$.

\paragraph{Model architecture.}
Like the coherence discriminator, this model also takes a form of dual encoder.
Given $(s_{k}, s_{k+1})$, $D_{\text{cohesion}}$ computes the cohesion score in three steps, as illustrated in Figure~\ref{dssm} (lower).
The first step is to obtain two sequences of word embedding to represent the two sentences. 
Then, a pair of source network $u(\mathord{\cdot})$ and target network $v(\mathord{\cdot})$ are utilized to encode both $s_{k}$ and $s_{k+1}$ into two low-dimensional continuous vectors. The two encoders $u(\mathord{\cdot})$ and $v(\mathord{\cdot})$ share the same architecture but do \textit{not} share parameters, i.e.,  $\theta_u \neq \theta_v$, and thus the $D_{\text{cohesion}}\left(s_{k},s_{k+1} \right)$ is \textit{not} symmetric. 
Finally, $D_{\text{cohesion}}\left(s_{k},s_{k+1} \right)$ is computed as the cosine similarity of the two vectors. 

Note that we use CNNs or RNNs to embed sentences in $D_{\text{cohesion}}$, which takes the word order in a sentence into consideration. This is different from the BOW embedding in the $D_{\text{coherence}}$ where the word order does not matter, because the word order indeed matters when determining the cohesion of two consecutive sentences. As an example from Table~\ref{sanity}, for the source sentence \textit{``Once you get there you are greeted by \underline{the staff}.''}, \textit{``\underline{They} explain everything to you.''} is a cohesive follow-up while \textit{``You explain everything to them.''} is not.

%$s_{k}=$ \textit{once you get there you are greeted by \underline{the staff}} and $s_{k+1}=$ \textit{\underline{they explain} everything to you} demonstrates a smooth flow between the sentences and an importance of word order. 

The parameters of $D_{\text{cohesion}}$, $\theta_u$ and $\theta_v$ are optimized using the same pairwise ranking loss. The positive pairs (a training minibatch) for $D_{\text{cohesion}}$ is obtained from (1) decomposing each paragraph $(S, T)$ in $\{(S_i,T_i)\}_{i=1}^B$ into pairs of consecutive sentences and (2) randomly selecting $B$ pairs as the positive (cohesive) pairs $\{(s_{k},s_{k+1})_i\}_{i=1}^B$. We construct negative (incohesive) pairs using the same methods as in the coherence discriminator.   

%$\{(s_{k},\widetilde{s}_{k+1,j})\}_{j=1}^{2B-1}$
%\paragraph{Constructing negative (incohesive) pairs for $D_{\text{cohesion}}$ by rotation and shuffling.}
\paragraph{Constructing negative (incohesive) pairs.}

% Given a training minibatch $\{(S_i,T_i)\}_{i=1}^B$, we construct $2*B-1$ negative pairs $\{(S_i,\widetilde{T_{i,j}})\}_{j=1}^{2B-1}$ for \textit{every} positive pair $(S_i, T_i)$ using three different methods, inspired by \citet{wieting2016iclr}. For notation simplicity, we omit the batch index $i$ in the rest of this section. For each positive pair $(S, T)$ in the batch:

We construct $2*B-1$ negative pairs $\{(s_{k},\widetilde{s}_{k+1,j})_i\}_{j=1}^{2B-1}$ for \textit{every} positive pair $(s_{k},s_{k+1})_i$ using three different methods and omit the minibatch index $i$ hereafter. For each positive pair $(s_{k},s_{k+1})$ in the minibatch:

% We construct negative samples for $D_{\text{cohesion}}$ from a training batch $\{(s_{k},s_{k+1})_i\}_{i=1}^B$ similar to how we construct negative samples for $D_{\text{coherence}}$ from a training batch $\{(S_i,T_i)\}_{i=1}^B$. For notation simplicity, we omit the batch index $i$ in the rest of this section. For each positive pair $(S, T)$ in the batch:
\begin{compactitem}
\item We mismatch sentence pairs to obtain $\{(s_{k},\widetilde{s}_{k+1,j})\}_{j=1}^{B-1}$. 
\item We shuffle words in $s_{k+1}$ to obtain $\widetilde{s}_{k+1}$. 
\item We combine the previous two methods and obtain additional pairs $\{(s_{k},\widetilde{s}_{k+1,j})\}_{j=1}^{B-1}$.
\end{compactitem}
In total, we obtain $2B-1$ negative pairs for each positive pair in the minibatch.
% mismatch pairs of sentences to obtain $\{(s_{k},s_{k+1}),(s_{k},\widetilde{s}_{k+1,j})\}_{j=1}^{B-1}$, and also shuffle words in $s_{k+1}$ to obtain $\widetilde{s}_{k+1}$. Thirdly, we combine the previous two steps, to obtain additional $\{(s_{k},s_{k+1}), (s_{k},\widetilde{s}_{k+1,j})\}_{j=1}^{B-1}$, totalling $2B-1$ negative samples for each of the sample in the batch.

\paragraph{Training using a pairwise ranking loss.} 

The parameters of $u(\mathord{\cdot})$ and $v(\mathord{\cdot})$ are optimized such that $D_{\text{cohesion}}(s_{k},s_{k+1}) >  D_{\text{cohesion}} (s_{k}, \widetilde{s}_{k+1,j})$ for any $j$. 
To achieve this, we propose to minimize the following pairwise ranking loss with margin $\delta$:
\begin{equation}\label{cohesion_loss_function} 
\resizebox{0.89\columnwidth}{!}{$
\begin{aligned}
	&L_{\text{cohesion}}(\theta_u,\theta_v) \coloneqq \max\Big(0, \delta - D_{\text{cohesion}}(s_{k},s_{k+1}) \\
	&+ \text{AVG}^{\lambda} \left( \{D_{\text{cohesion}} (s_{k}, \widetilde{s}_{k+1,j})\}_{j=1}^{2B-1} \right) \Big).
\end{aligned}$}
\end{equation}

% Similar to training $D_{\text{coherence}}$, we define 
% \begin{equation}
% \label{modified_cohesion_margin}
% % \resizebox{0.89\columnwidth}{!}{$
% \begin{aligned}
% & \Delta^{\circ}(\theta_u,\theta_v) \coloneqq  D_{\text{cohesion}}(s_{k},s_{k+1}) \\ & - \text{AVG}^{\lambda} \left( \{D_{\text{cohesion}} (s_{k}, \widetilde{s}_{k+1,j})\}_{j=1}^{2B-1} \right)
% % \end{equation}
% %\]
% \end{aligned}%$}
% \end{equation}
% and, given a training batch $\{(s_{k},s_{k+1})_i\}_{i=1}^B$ and their negative samples $\{(s_{k},\widetilde{s}_{k+1,j})\}_{j=1}^{2B-1}$, we optimize parameters $(\theta_u,\theta_v)$ by minimizing the following loss. 
% \begin{equation}
% \label{cohesion_loss_function}
% \resizebox{0.89\columnwidth}{!}{$
% L_{\text{cohesion}}(\mathord{\cdot}) \coloneqq \frac{\displaystyle 1}{\displaystyle B} \sum\limits_{i=1}^B \max\Big(0, \delta-\Delta^{\circ}(\theta_u,\theta_v)^{(i)}\Big)
% $}
% \end{equation}

We leave the training details and hyper-parameter configurations to Section~\ref{implementation}.

% $s_{i} \coloneqq \left[ s_{i,1}, s_{i,2},..., s_{i,n} \right]$ and $s_{i+1} \coloneqq \left[ s_{i+1,1}, s_{i+1,2},..., s_{i+1,m} \right]$, where $s_{i,k}$ denotes the $k$-th word in sentence $i$.

% Given these signals, we use our proposed variant of the policy gradient, negative-critical sequence training, to apply parameter updates. %We discuss the details in the next section.

\begin{table*}[t!]
\centering
\resizebox{\textwidth}{!}{%
\begin{tabular}{lcl}
\textbf{source}   & \textbf{cohesion}   & \multicolumn{1}{c}{\textbf{coherence}}  \\ \hline
{\color[HTML]{CB0000} this hotel was unbelievably overpriced .}& {\color[HTML]{343434} 0.0002}&\\
\begin{tabular}[c]{@{}l@{}}we were looking for something cheaper but thought we would at least \\ be staying in a decent hotel having paid that much when booking .\end{tabular}& {\color[HTML]{343434} 0.0411}&\\
it wasn t clear when booking that we would have to {\color[HTML]{CB0000} share a bathroom .}& {\color[HTML]{343434} 0.0084}&\\
there was one shower for the whole floor which was tiny and unclean .& {\color[HTML]{343434} 0.0054}&\\
the room was old and lacking in facilities .&&\\
&&\\
\textbf{target}&&\\ \hline
the beds were very uncomfortable and the linen was very old .& {\color[HTML]{343434} 0.0768}& \multicolumn{1}{c}{{\color[HTML]{343434} }}\\
breakfast was ok , but the staff were incompetent .& {\color[HTML]{343434} 0.0591}& \multicolumn{1}{c}{{\color[HTML]{343434} }}\\
on our last day they were too lazy to clean our table and never bothered taking our order .& {\color[HTML]{343434} -0.0097}& \multicolumn{1}{c}{{\color[HTML]{343434} }}\\
we had to leave having had no breakfast , as we ran out of time .& {\color[HTML]{343434} 0.0457}& \multicolumn{1}{c}{{\color[HTML]{343434} }}\\
they saw us get up and leave and didn t even apologise for the appalling lack of service .&& \multicolumn{1}{c}{\multirow{-5}{*}{{\color[HTML]{343434} \textbf{+0.3735}}}} \\
&&\\
\textbf{negative target}&&\\ \hline
the staff recommended great restaurants with very reasonable prices within walking distance .& {\color[HTML]{343434} 0.0514}& \multicolumn{1}{c}{{\color[HTML]{CB0000} }}\\
the paris hop on bus stops nearby .& {\color[HTML]{343434} 0.0798}& \multicolumn{1}{c}{{\color[HTML]{CB0000} }}\\
the gare l est is within 3 blocks .& {\color[HTML]{343434} -0.0156}& \multicolumn{1}{c}{{\color[HTML]{CB0000} }}\\
\begin{tabular}[c]{@{}l@{}}we paid 75 euro per nite excluding breakfast but paid for breakfast one day and found it {\color[HTML]{CB0000} very} \\ {\color[HTML]{CB0000} good and reasonably priced .}\end{tabular} & {\color[HTML]{343434} 0.0082}& \multicolumn{1}{c}{{\color[HTML]{CB0000} }}\\
the rooms are clean and {\color[HTML]{CB0000} bathrooms ensuite .}&& \multicolumn{1}{c}{\multirow{-5}{*}{{\color[HTML]{CB0000} \textbf{-0.2001}}}} \\
& \multicolumn{1}{l}{}&\\ \hline
\textbf{more examples of cohesion}& \multicolumn{1}{l}{}&\\ \hline
once you get there you are greeted by the staff .&&\\
they explain everything to you , and in english , not the best , but good enough .& \multirow{-2}{*}{0.1004}&\\ \hline
the coffee was even good for a coffee snob like myself .& {\color[HTML]{CB0000} }&\\
the hotel is much smaller than i thought and only has six floors .& \multirow{-2}{*}{{\color[HTML]{CB0000} -0.1103}} &\\ \hline
the only negative was the curtain in the bathroom .&&\\
\begin{tabular}[c]{@{}l@{}}it was very shear and we felt that people in the building across the street could look \\ right in at night .\end{tabular}& \multirow{-2}{*}{0.0787}&\\ \hline
the beer at the lobby bar was stale .& {\color[HTML]{CB0000} }&\\
there are many friendly cats on the grounds .& \multirow{-2}{*}{{\color[HTML]{CB0000} -0.0830}} &\\ \hline
\end{tabular}%
}
\caption{Coherence and cohesion rewards on test data. The cohesion reward at the end of each line is computed with its next sentence. This is an example of contradiction and inconsistent sentiment, suggestive of incoherence. We append more examples with extreme cohesion rewards.}
\label{sanity}
\end{table*}

\section{Negative-Critical Sequence Training for Long-form Text Generation}
\label{ncst}
\subsection{Long-form text generator: \texorpdfstring{$G$}}
The generator $G$ is an attention-based bidirectional sequence-to-sequence model \citep{DBLP:journals/corr/BahdanauCB14} and is pre-trained by maximizing the log likelihood on training data, which we denote as $G_{\text{MLE}}$. However, long-form texts generated using $G_{\text{MLE}}$ often do not meet our high coherence and cohesion standards.

We propose to use the two pre-trained discriminators, \texorpdfstring{$D_{\text{coherence}}$}{Lg} and \texorpdfstring{$D_{\text{cohesion}}$}{Lg}, to modify the text generation behavior of $G_{\text{MLE}}$. The scores from the discriminators are used as reward (or penalty) signals to adjust the parameters of $G_{\text{MLE}}$ using a variant of policy gradient, called \emph{negative-critical sequence training}, which we propose for our task and describe in details in the next subsection.

\subsection{Negative-critical sequence training}
For an arbitrary pair of $S$ and $T_{gen}$, where $T_{gen}$ is the generator's output conditioned on $S$, we compute the coherence and cohesion scores by calling $D_{\text{coherence}}$ and $D_{\text{cohesion}}$. Since each generated text consists of multiple sentences, the overall cohesion score is computed as the mean of all the consecutive sentence pairs, $(s_{k},s_{k+1}) \subset \left[S_{-1}, T_{gen}\right]$, where $S_{-1}$ is the last sentence from the source.

These scalar scores, however, are not interpretable since the discriminators are trained by optimizing a pairwise ranking loss. Instead, the differences between positive pair scores and the maximal or average negative pair scores provide insights of how well the models distinguish between the positive and the negative pairs. 

This difference relates to reward with baseline in actor-critic methods \citep{witten1977adaptive,barto1983neuronlike,Williams:1992:SSG:139611.139614,Sutton:1999:PGM:3009657.3009806} that typically require a separate critic function as a baseline. In NLP, we have observed similar practices by \citet{DBLP:journals/corr/RanzatoCAZ15}, \citet{bahdanau2016actor}, and \citet{Nguyen2017ReinforcementLF}. \citet{Rennie2017SelfCriticalST} proposed a method that avoids learning a separate critic. Similarly, our method does not require learning a separate critic since this margin is a form of reward minus baseline. Specifically, we define the reward functions with baselines as:
\begin{equation}
\begin{aligned}
R_{\text{coherence}}(S,T_{gen})  & \coloneqq D_{\text{coherence}}(S,T_{gen}) \\ 
&- {\mathbb{E}_{\widetilde{T}}} \left[D_{\text{coherence}}(S,\widetilde{T})\right] 
\end{aligned}
\end{equation}
\begin{equation}
\begin{aligned}
& R_{\text{cohesion}}(\left[S_{-1}, T_{gen}\right]) \coloneqq 
\\
&\;\;\; \frac{1}{|T_{gen}|} \sum \limits_{\substack{ (s_{k},s_{k+1}) \\ \subset \left[S_{-1}, T_{gen}\right]}} D_{\text{cohesion}}(s_{k},s_{k+1})
\\
&\;\;\; -{\mathbb{E}_{\widetilde{s}_{k+1} \mathrel{\Big|} \substack{(s_{k},s_{k+1}) \\ \subset \left[S, T\right]}}} \Big[D_{\text{cohesion}}(s_{k},\widetilde{s}_{k+1}) \Big]
\end{aligned}
\end{equation}
where $|T_{gen}|$ denotes the number of sentences in $T_{gen}$, 
% $\tilde{\cdot}$ denotes a negative sample for a given source condition, 
and $\mathbb{E}_{\widetilde{T}}$ ( and $\mathbb{E}_{\widetilde{s}_{k+1}}$) are computed by averaging over an ensemble of negative pairs. 

Notice that this reward resembles the ranking loss we use to train our discriminators, except that our baseline is the mean score (instead of the weighted mean) over negative pairs. The rationale for this difference is that: because the best artificially constructed negative sample may be a \textit{formidably} good sample, the maximal or the weighted mean can in fact be noisy as a baseline and thus introduce noise in rewards.
% Actor-critic methods \citep{barto1983neuronlike,witten1977adaptive} based on neural networks typically require learning a separate critic network to estimate the expected future reward as a \textit{baseline}, which in many cases is a difficult task by itself. In NLP, we have observed similar practices and challenges by \citet{DBLP:journals/corr/RanzatoCAZ15}, \citet{bahdanau2016actor}, and \citet{Nguyen2017ReinforcementLF}. 
% Recently, \citet{Rennie2017SelfCriticalST} proposed an effective self-critical sequence training method that avoids learning a separate critic network. 
% Similarly, the proposed negative-critical sequence training method does not require learning a separate critic network. Instead, we directly use the scores of negative samples assigned by the discriminators as the \textit{baseline}. In what follows, we describe the method in detail.
To alleviate such noise, we use the \textit{mean discriminator score} of negative pairs as the baseline, and this turns out to be an empirically better alternative. Then we use policy gradient to maximize a weighted sum of the coherence and cohesion rewards. %For illustrative purposes, we equally weigh them for updating our policy, i.e., the generator $G$.

\begin{table*}[t!]
\centering
\resizebox{\textwidth}{!}{%
\begin{tabular}{@{}clcccccclccc@{}}
\cmidrule(r){1-5} \cmidrule(l){8-12}
\multicolumn{2}{c}{\textbf{TripAdvisor}} & \multicolumn{3}{c}{\textbf{Target Sentences Retrieval}} &  &  & \multicolumn{2}{c}{\textbf{Yelp}} & \multicolumn{3}{c}{\textbf{Target Sentences Retrieval}} \\ \cmidrule(r){1-5} \cmidrule(l){8-12} 
\textbf{Discriminators} & \textbf{Encoding} & \textbf{R@1} & \textbf{R@5} & \textbf{R@10} &  &  & \textbf{Discriminators} & \textbf{Encoding} & \textbf{R@1} & \textbf{R@5} & \textbf{R@10} \\ \cmidrule(r){1-5} \cmidrule(l){8-12} 
\multirow{2}{*}{$D_{\text{coherence}}$} & $\text{Conv}^{512}_{2,3,4,5}$ & 0.18 & 0.43 & 0.60 &  &  & \multirow{2}{*}{$D_{\text{coherence}}$} & $\text{Conv}^{512}_{2,3,4,5}$ & 0.33 & 0.61 & 0.74 \\ \cmidrule(lr){2-5} \cmidrule(l){9-12} 
 & $\text{GRU}^{1024}_{\text{1-layer, bi-dir.}}$ & \textbf{0.26} & \textbf{0.50} & \textbf{0.65} &  &  &  & $\text{GRU}^{1024}_{\text{1-layer, bi-dir.}}$ & \textbf{0.39} & \textbf{0.68} & \textbf{0.81} \\ \cmidrule(r){1-5} \cmidrule(l){8-12} 
\multirow{2}{*}{$D_{\text{cohesion}}$} & $\text{Conv}^{512}_{3,4,5,6}$ & \textbf{0.12} & \textbf{0.28} & \textbf{0.43} &  &  & \multirow{2}{*}{$D_{\text{cohesion}}$} & $\text{Conv}^{512}_{3,4,5,6}$ & \textbf{0.14} & \textbf{0.33} & \textbf{0.47} \\ \cmidrule(lr){2-5} \cmidrule(l){9-12} 
 & $\text{GRU}^{1024}_{\text{1-layer, bi-dir.}}$ & 0.11 & 0.21 & 0.33 &  &  &  & $\text{GRU}^{1024}_{\text{1-layer, bi-dir.}}$ & 0.11 & 0.26 & 0.39 \\ \cmidrule(r){1-5} \cmidrule(l){8-12} 
\end{tabular}%
}
\caption{Retrieval ratios for coherence and cohesion discriminators from a collection of 100 negative candidates from the test data. The reported numbers are the averages over 20 evaluations. Notations: $\text{Conv}^{512}_{2,3,4,5}$ is a convolutional input encoder with filter sizes 2, 3, 4, and 5, and there are 512 filters for each filter size. $\text{GRU}^{1024}_{\text{1-layer, bi-dir.}}$ is a 1-layered bi-directional GRU input encoder with hidden size 1024. We experimented different configurations for both encoder types, and selected the best performing models for the negative-critical sequence training step. %For a detailed description, see Section~\ref{implementation}.
}
\label{retrieval}
\end{table*}

\begin{table*}[t!]
\centering
\resizebox{\textwidth}{!}{%
\begin{tabular}{@{}ll@{}}
\toprule
\textbf{source sentences}  & \begin{tabular}[c]{@{}l@{}}the hotel inglaterra delivered as promised . the staff was welcoming and spoke good english . the cleaning staff did a \\ very good job every day . the rooms were spotless and very modern . the bathroom was large and had a very nice shower \\ , and there were two generously sized bath towels that were twice the size of normal towels .\end{tabular}  
\\ \midrule
\textbf{$G_{\text{MLE}}$}  & \begin{tabular}[c]{@{}l@{}}\textcolor{red}{the breakfast in the morning was delicious and very good .} it was the only hotel where i slept very well . the staff was\\ very helpful in late afternoon or late times . \textcolor{red}{the breakfast was adequate , with a decent range of cereals , fruit , and} \\ \textcolor{red}{fruits .} there is also free use of the coffee in the reception area .\end{tabular} 
\\ \midrule
\textbf{$G_{\text{MLE+RL(coherence, cohesion)}}$}  & \begin{tabular}[c]{@{}l@{}}the breakfast was plentiful including fresh breads and cooked to order . the location was fantastic . it is in the north \\ of the marina and in a very short distance . the marina has a small swimming pool with sitting area and a small gym .\\ they are very popular and guests have an evening reception which is very nice .\end{tabular}
\\ \bottomrule
\end{tabular}%
}
\caption{Sample generations from our MLE-trained baseline model, $G_{\text{MLE}}$, and our discriminator-guided model $G_{\text{MLE+RL(coherence, cohesion)}}$. The \textcolor{red}{red texts} highlight a common problem in $G_{\text{MLE}}$ - it exhibits a repetition, and an inconsistent opinion as a review. In contrast, our discriminator-guided model is able to generate a more interesting, and sentiment-consistent continuation.}
\label{sample}
\end{table*}

\begin{table*}[t!]%[ht!]
\centering
\resizebox{\textwidth}{!}{%
\begin{tabular}{clcccccccccc}
\hline
\multicolumn{1}{c|}{\multirow{5}{*}{\textbf{TripAdvisor}}} & \multicolumn{1}{c}{\textbf{Model}} & \textbf{NLL} & \textbf{PPL} & \textbf{BLEU-3} & \textbf{BLEU-4} & \textbf{BLEU-5} & \multicolumn{1}{l}{\textbf{\begin{tabular}[c]{@{}l@{}}intra-\\ unique-1\end{tabular}}} & \multicolumn{1}{l}{\textbf{\begin{tabular}[c]{@{}l@{}}intra-\\ unique-2\end{tabular}}} & \multicolumn{1}{l}{\textbf{\begin{tabular}[c]{@{}l@{}}inter-\\ unique-2\end{tabular}}} & \multicolumn{1}{l}{\textbf{\begin{tabular}[c]{@{}l@{}}inter-\\ unique-3\end{tabular}}} & \multicolumn{1}{l}{\textbf{\begin{tabular}[c]{@{}l@{}}length \\ ratio\end{tabular}}} \\ \cline{2-12} 
\multicolumn{1}{c|}{} & $G_{\text{MLE}}$ (baseline) & 0.86 & 2.36 & 0.38 & 0.19 & 0.08 & 0.66 & 0.93 & 0.40 & 0.72 & 1.08 \\ \cline{2-12} 
\multicolumn{1}{c|}{} & $G_{\text{MLE +RL(cohesion)}}$ & \textbf{0.77} & \textbf{2.18} & \textbf{0.46} & \textbf{0.27} & \textbf{0.14} & 0.64 & 0.94 & 0.38 & 0.71 & 0.97 \\ \cline{2-12} 
\multicolumn{1}{c|}{} & $G_{\text{MLE+RL(coherence)}}$ & 0.80 & 2.24 & 0.44 & 0.25 & 0.12 & 0.64 & 0.94 & 0.39 & 0.72 & 1.06 \\ \cline{2-12} 
\multicolumn{1}{c|}{} & $G_{\text{MLE+RL(coherence, cohesion)}}$ & 0.80 & 2.25 & 0.44 & 0.24 & 0.12 & 0.65 & 0.94 & 0.40 & 0.72 & 1.02 \\ \hline
 &  & \multicolumn{1}{l}{} & \multicolumn{1}{l}{} & \multicolumn{1}{l}{} & \multicolumn{1}{l}{} & \multicolumn{1}{l}{} & \multicolumn{1}{l}{} & \multicolumn{1}{l}{} & \multicolumn{1}{l}{} & \multicolumn{1}{l}{} & \multicolumn{1}{l}{} \\ \hline
\multicolumn{1}{c|}{\multirow{5}{*}{\textbf{Yelp}}} & \multicolumn{1}{c}{\textbf{Model}} & \multicolumn{1}{l}{\textbf{NLL}} & \multicolumn{1}{l}{\textbf{PPL}} & \multicolumn{1}{l}{\textbf{BLEU-3}} & \multicolumn{1}{l}{\textbf{BLEU-4}} & \multicolumn{1}{l}{\textbf{BLEU-5}} & \multicolumn{1}{l}{\textbf{\begin{tabular}[c]{@{}l@{}}intra-\\ unique-1\end{tabular}}} & \multicolumn{1}{l}{\textbf{\begin{tabular}[c]{@{}l@{}}intra-\\ unique-2\end{tabular}}} & \multicolumn{1}{l}{\textbf{\begin{tabular}[c]{@{}l@{}}inter-\\ unique-2\end{tabular}}} & \multicolumn{1}{l}{\textbf{\begin{tabular}[c]{@{}l@{}}inter-\\ unique-3\end{tabular}}} & \multicolumn{1}{l}{\textbf{\begin{tabular}[c]{@{}l@{}}length \\ ratio\end{tabular}}} \\ \cline{2-12} 
\multicolumn{1}{c|}{} & $G_{\text{MLE}}$ (baseline) & 1.32 & 3.84 & 0.37 & 0.17 & 0.07 & 0.68 & 0.95 & 0.54 & 0.86 & 1.07 \\ \cline{2-12} 
\multicolumn{1}{c|}{} & $G_{\text{MLE+RL(cohesion)}}$ & 1.26 & 3.65 & \textbf{0.45} & \textbf{0.23} & \textbf{0.11} & 0.68 & 0.95 & 0.53 & 0.85 & 1.05 \\ \cline{2-12} 
\multicolumn{1}{c|}{} & $G_{\text{MLE+RL(coherence)}}$ & \textbf{1.24} & \textbf{3.56} & 0.45 & 0.23 & 0.11 & 0.69 & 0.95 & 0.55 & 0.87 & 1.00 \\ \cline{2-12} 
\multicolumn{1}{c|}{} & $G_{\text{MLE+RL(coherence, cohesion)}}$ & 1.25 & 3.59 & 0.43 & 0.22 & 0.11 & 0.69 & 0.95 & 0.56 & 0.88 & 1.05 \\ \hline
\end{tabular}%
}
\caption{An ablation study with automated evaluation metric scores: NLL, PPL, BLEU-$n$, intra/inter-unique-$n$, along with the length ratio with the length of corresponding true target sentences as 1. Significant numbers are highlighted in \textbf{bold} before rounding.}
\label{modelstats}
\end{table*}
\section{Experiments}
\label{experiments}

In this section, we detail the training and evaluation of $D_{\text{coherence}}$, $D_{\text{cohesion}}$, the baseline generator $G_{\text{MLE}}$, and the RL-tuned generators $G_{\text{MLE+RL(cohesion)}}$, $G_{\text{MLE+RL(coherence)}}$, and $G_{\text{MLE+RL(coherence, cohesion)}}$. We show that, by using feedback from the discriminators, the quality of the generated texts is significantly improved. See Table~\ref{sample} for a sample comparison.

\subsection{Dataset}
We use the TripAdvisor hotel English reviews dataset collected by \citet{Wang2010LatentAR} and the Yelp English reviews dataset\footnote{https://www.yelp.com/dataset}. We use only the subsets of the two datasets that satisfy the following two conditions: (1) a review must have at least 10 sentences, and (2) each sentence has from 5 to 30 words. This yields roughly 60,000 TripAdvisor reviews and 220,000 Yelp reviews, split into $\left[0.8, 0.1, 0.1 \right]$ ratio for train/dev/test sets.

We merge the source and target vocabularies, and limit it to the top 50,000 frequent words, excluding special tokens. For each review, we use the first five sentences as the input $S$ to $G$, and the next five sentences as the target output $T$ from $G$. 

\subsection{Implementation details} 
\label{implementation}
% \textbf{Implementation details}.

\paragraph{Baseline $G_{\text{MLE}}$.} 
$G_{\text{MLE}}$ takes individual words as inputs and embeds into a pre-trained GloVe 300-dimensional word vectors. This embedding layer is fixed throughout training. 
$G_{\text{MLE}}$ uses a two-layered GRU and hidden size of 1024 for both encoder and decoder. During optimization using Adam \citep{Kingma2014AdamAM}, we set the learning rate to $2\text{e-}4$ and clip the gradient's L2-norm to 1.0. We initially train $G_{\text{MLE}}$ for 60 epochs on the TripAdvisor data and 30 epochs on the Yelp data.%We initially train $G_{\text{MLE}}$ by maximum likelihood estimation (MLE) on the training data that consist of positive samples for 60 epochs on the TripAdvisor data and 30 epochs on the Yelp dataset. % These are our benchmark baseline models.

\begin{table*}[t!]
\small
\centering
\begin{tabular}{r r | r| r l  l r r | r | r | l}
\cmidrule[\heavyrulewidth]{1-11}
 \multicolumn{5}{c}{Cohesion} & & \multicolumn{5}{c}{Coherence}\\
\cmidrule[\heavyrulewidth]{1-5} \cmidrule[\heavyrulewidth]{7-11}
\multicolumn{5}{c}{{\it Human judges preferred:}} & & \multicolumn{5}{c}{{\it Human judges preferred:}} \\
\cmidrule[\heavyrulewidth]{1-5} \cmidrule[\heavyrulewidth]{7-11}
\multicolumn{2}{c|}{Our Method} & Neutral & \multicolumn{2}{c}{Comparison} & & \multicolumn{2}{c|}{Our Method} & Neutral & \multicolumn{2}{c}{Comparison} \\ 
\cmidrule[\heavyrulewidth]{1-5} \cmidrule[\heavyrulewidth]{7-11}
\cmidrule{1-5} \cmidrule{7-11}
$G_{\text{MLE+RL}}$ & \bf{36.41}\% & 33.57\% & 30.50\% & $G_{\text{MLE}}$ & & $G_{\text{MLE+RL}}$ & \bf{37.23}\% & 31.44\% & 31.80\% & $G_{\text{MLE}}$\\
$G_{\text{MLE+RL}}$ & 29.91\% & 30.85\% & \bf{39.24}\% & Human & & $G_{\text{MLE+RL}}$ & 28.96\% & 31.32\% & \bf{39.72}\% & Human \\
	\cmidrule[\heavyrulewidth]{1-11}
	\end{tabular}
\caption{Results of {\bf Human Evaluation} showing preferences (\%) for our model $G_{\text{MLE+RL(coherence, cohesion)}}$ vis-a-vis the baseline $G_{\text{MLE}}$ after adjustment for spamming. $G_{\text{MLE+RL(coherence, cohesion)}}$ is preferred over $G_{\text{MLE}}$. For simplicity, the 5-point Likert scale has been collapsed to a 3-point scale. 
%*Differences in mean preferences are statistically significant (p  $\leq$ 0.0001). 
See the Appendix for further details of distributions. }\label{tab:humanevalresults} 
\end{table*}

\paragraph{Discriminators.} 
For the CNN-based encoder, the convolutional layer consists of filters of sizes 2, 3, 4, and 5 for $D_{\text{coherence}}$ (3, 4, 5, and 6 for $D_{\text{cohesion}}$), each with 512 filters. Each convolution filter is followed by a $\tanh$ activation. Then, we max-pool in time and append a fully connected layer to generate a feature vector of dimension 512, followed by a batch normalization layer and a $\tanh$ activation. 
For the RNN-based encoder, we use a 1-layered bi-directional GRU, concatenate the final hidden states at both ends, and append the same remaining layers. 

Both discriminators use the pre-trained GloVe word embedding vectors\footnote{The vector dimension can be different from that of $G$. The differences were marginal for sizes 50, 100, and 300. For results shown in this paper, we used the same dimension of size 300.}, which are fixed during the training. We use an Adam optimizer with a learning rate of $1\text{e-}5$. We fix $\lambda=2$ and $\delta=0.2$ in equations~(\ref{coherence_loss_function}) and~(\ref{cohesion_loss_function}).\footnote{We performed a coarse grid search over the values of $\lambda$ and $\delta$ and these values for the hyper-parameters pair resulted in fast convergence to high recall scores on the dev dataset.} We train both discriminators for 50 epochs and choose the models with the best R$@$1 scores on the validation dataset.

\paragraph{Model $G_{\text{MLE+RL}}$.}
In the fine-tuning stage, we use the negative-critical sequence training method, as described in Section~\ref{ncst}, up to 5 epochs, with a learning rate of $1\text{e-}5$. 
We equally weight the coherence and cohesion rewards, $\frac{1}{2} R_{\text{coherence}}(S,T_{gen}) + \frac{1}{2} R_{\text{cohesion}}(\left[S_{-1}, T_{gen}\right])$. 
We also continue the supervised learning of $G$ to constrain the policy search within a space that represents the sentences that are likely to be grammatically plausible, similar to \citet{wu2016google,DBLP:journals/corr/PaulusXS17,lewis2017deal}. 
For all the generations from $G_{\text{MLE}}$ and $G_{\text{MLE+RL}}$, we use the simple greedy decoding method because we do not observe any significant difference when switching to beam search.

\subsection{Results}

\paragraph{Evaluating \texorpdfstring{$D_{\text{coherence}}$}{Lg} and \texorpdfstring{$D_{\text{cohesion}}$}{Lg}.} 
Since the discriminators are implemented as pairwise rankers, we employ the metrics commonly used in information retrieval for evaluation, i.e., recall at \textit{K} (R$@$\textit{K}), which is defined as the fraction of correctly identifying an item in the TOP-\textit{K} retrieved list \citep{baeza1999modern}. 
We present the retrieval results in Table~\ref{retrieval}. 
To help readers understand the roles of $D_{\text{coherence}}$ and $D_{\text{cohesion}}$, we present examples of positive and negative pairs and their rewards in Table~\ref{sanity}. 

\paragraph{\bf Automatic evaluation of  \texorpdfstring{$G$}{Lg}.} 
It is widely known that there is no perfect automated metric to evaluate text generators. Nevertheless, we report the scores of widely used metrics, including negative log-likelihood (NLL), perplexity (PPL), BLEU and the proportion of unique $n$-grams within a single generation (intra-unique-$n$), and across generations (inter-unique-$n$), as in \citet{DBLP:journals/corr/abs-1805-12352}. 
Results in Table~\ref{modelstats} show that our discriminators significantly improve BLEU scores, NLL and PPL, with marginal difference in diversity.

\paragraph{Human evaluation of \texorpdfstring{$G$}{Lg}.} Coherence and cohesion of a text cannot be easily measured using standard automated metrics. Thus, we perform crowd-sourced human evaluation.
%on the generation quality with respect to coherence and cohesion. 
We randomly selected 200 samples from the TripAdvisor dataset, including corresponding generated output from the baseline $G_{\text{MLE}}$ and our model $G_{\text{MLE+RL}}$.
For comparison, we pair systems as $\left(\textit{Human} \leftrightarrow G_{\text{MLE+RL}}\right)$ and $\left(G_{\text{MLE+RL}} \leftrightarrow G_{\text{MLE}}\right)$. 

The outputs of these system pairs are presented in random order and each is ranked in terms of coherence and cohesion using a five-point Likert scale by human judges. Initially, we hired 7 judges to judge each pair. We identified a group of poor judges (probable spammers) who choose $G_{\text{MLE+RL}}$ over the \textit{Human} more than 40\% of the time, and eliminated them from the judge pool. 
Table~\ref{tab:humanevalresults} reports the final scores in terms of percentages of the total remaining judgments.

% \JG{You need to summarize the results by saying something like following: ``Our automatic evaluation results are consistent with human evaluation results that (1) the proposed method does generate more locally and globally consistent texts thanks to the use of the cohesion and coherent discriminators... (2) the Negative-Critical Training method turns out to be very effective in training text generators via RL (not sure if you have any empirical result to support this)...''}

% \subsection{Discussion}
% % \textbf{Discussion}.
% In Table~\ref{retrieval}, notice that for RNNs outperform CNNs for coherence models, and CNNs outperform RNNs for cohesion models. One explanation is that RNNs are effective in encoding a sequential input yet exhibit drawbacks when encoding into hidden states at both ends of a \textit{long} input, otherwise well-known as a long-range dependency problem. 

% While reinforcing coherence and cohesion properties in text generation through surrogate models is an important research direction, we consider our results to be preliminary, and our experiment results allude to room for improvement, such as recall scores. This is because our methods to construct negative samples from an unlabelled dataset are not thorough. For example, a randomly mismatched sentence that follows a given sentence may actually be a valid continuation. In this work, we overlook this problem since our proposed schemes are shown to be effective in modeling coherence and cohesion.

\section{Conclusion}

This paper proposes a neural approach to explicitly modeling cross-sentence linguistic properties, coherence and cohesion, for long-form text generation. The coherence discriminator $D_{\text{coherence}}$ provides a macro-level view on structuring a paragraph. The cohesion discriminator $D_{\text{cohesion}}$ provides a micro-level view on local connectivity between neighboring sentences.  
The pre-trained discriminators are used to score the generated texts and artificially constructed negative pair scores are used to form baselines for the policy gradient, which we call negative-critical sequence training, to train neural language models. 
% The positive pair scores computed by these discriminators are used as rewards, and the negative pairs are used to form a baseline in policy gradient, which we call negative-critical sequence training, to train neural language models. 

On two long-form text generation tasks, human evaluation results are consistent with automatic evaluation results, which together demonstrate that our proposed method generates more locally and globally consistent texts with the help of the discriminators. %on two long-form text generation tasks show that our coherence and cohesion models improve upon the strong baseline, an attention-based bidirectional MLE-trained sequence-to-sequence model. 

Despite the encouraging initial results, we only scratched the surface of the problem. The proposed method is yet to be significantly improved to meet the ultimate goal of generating meaningful and logical long-form texts.

% In this paper, we propose to measure coherence and cohesion of a text via models parametrized by neural networks trained through simple mechanisms. The coherence discriminator $D_{\text{coherence}}$ provides a macro-level view on structuring a paragraph. It assesses how likely two text chunks form a coherent paragraph, using sentence-level features. On the other hand, the cohesion discriminator $D_{\text{cohesion}}$ provides a micro-level view on local connectivity between neighboring sentences. It assesses how cohesive two consecutive sentences are, using word-level features. The scores computed by these discriminators are used as reward signals, and negative samples are used as baseline in policy gradient, called negative-critical sequence training, to train neural language models. 
% Empirical results from human evaluation and automatic metrics on two long-form text generation tasks show that our coherence and cohesion models improve upon the strong baseline, an attention-based bidirectional MLE-trained sequence-to-sequence model.

% Future work will focus on casting the long-form text generation task using the GANs framework. In this framework, the coherence and cohesion discriminators are modified against model-generated texts, and in turn, provide signals to learn neural language models.

\bibliography{naaclhlt2019}
\bibliographystyle{acl_natbib}

\begin{table*}[t!]
\small
\centering
\begin{tabular}{r r | r| r l  l r r | r | r | l}
\cmidrule[\heavyrulewidth]{1-11}
 \multicolumn{5}{c}{Cohesion} & & \multicolumn{5}{c}{Coherence}\\
\cmidrule[\heavyrulewidth]{1-5} \cmidrule[\heavyrulewidth]{7-11}
\multicolumn{5}{c}{{\it Human judges preferred:}} & & \multicolumn{5}{c}{{\it Human judges preferred:}} \\
\cmidrule[\heavyrulewidth]{1-5} \cmidrule[\heavyrulewidth]{7-11}
\multicolumn{2}{c|}{Our Method} & Neutral & \multicolumn{2}{c}{Comparison} & & \multicolumn{2}{c|}{Our Method} & Neutral & \multicolumn{2}{c}{Comparison} \\ 
\cmidrule[\heavyrulewidth]{1-5} \cmidrule[\heavyrulewidth]{7-11}
\cmidrule{1-5} \cmidrule{7-11}
$G_{\text{MLE+RL}}$ & 36.25\% & 26.62\% & \bf{37.13}\% & $G_{\text{MLE}}$ & & $G_{\text{MLE+RL}}$ & \bf{39.25}\% & 23.12\% & 37.63\% & $G_{\text{MLE}}$\\
$G_{\text{MLE+RL}}$ & 34.25\% & 23.63\% & \bf{42.12}\% & Human & & $G_{\text{MLE+RL}}$ & 35.63\% & 21.50\% & \bf{42.87}\% & Human \\
	\cmidrule[\heavyrulewidth]{1-11}
	\end{tabular}
\caption{Results of {\bf Human Evaluation} showing preferences (\%) for our model $G_{\text{MLE+RL(coherence, cohesion)}}$ vis-a-vis the baseline $G_{\text{MLE}}$ \emph{before} adjustment for spamming. For simplicity, the 5-point Likert scale has been collapsed to a 3-point scale.}\label{tab:humanevalresults_unadjusted} 
\end{table*}

\newpage
\appendix

\section{Human evaluation un-adjusted scores}
\label{sec:humaneval}
Crowd-sourced evaluation can be noisy because there may be human judges who do not take the task seriously, and rather randomly and/or deliberately choose options that prevent us from drawing accurate conclusions. Therefore, we removed crowd-sourced judges who chose $G_{\text{MLE+RL}}$ over the \textit{Human} more than 40\% of the time, which threshold value we considered appropriate to identify poor judges (probable spammers). In Table~\ref{tab:humanevalresults_unadjusted}, we present the un-adjusted results \emph{before} accounting for the poor judges.

\section{Sparse end-of-sequence rewards}
\label{sec:sparsereward}
Sequence-level rewards are available upon a completed generation, so they are sparse signals for the generator. In practice, sparse end-of-sequence rewards entail a noisy training, yet would want the learning generalize to the test data. We observed that, for our particular task, most noises were caused by exploration, and the learning generalized to the test data, as confirmed via both human and automatic evaluation results. Thus, reward shaping was unnecessary, unlike previous works \citep{li2017adversarial,yang2018unsupervised} that further provided signals for partially generated sequences.

\end{document}